# Interpretability of machine learning based prediction models in healthcare

(Advanced review)

**Gregor Stiglic[1], Primoz Kocbek[1], Nino Fijacko[1], Marinka Zitnik[2], Katrien Verbert[3], Leona Cilar[1]**

Affiliations:
1. University of Maribor, Slovenia
2. Harvard University, Boston, MA, USA
3. KU Leuven, Leuven, Belgium

**Abstract**
There is a need of ensuring machine learning models that are interpretable. Higher interpretability of the model means easier comprehension and explanation of future predictions for end-users. Further, interpretable machine learning models allow healthcare experts to make reasonable and data-driven decisions to provide personalized decisions that can ultimately lead to higher quality of service in healthcare. Generally, we can classify interpretability approaches in two groups where the first focuses on personalized interpretation (local interpretability) while the second summarizes prediction models on a population level (global interpretability). Alternatively, we can group interpretability methods into model-specific techniques, which are designed to interpret predictions generated by a specific model, such as a neural network, and model-agnostic approaches, which provide easy-to-understand explanations of predictions made by any machine learning model. Here, we give an overview of interpretability approaches and provide examples of practical interpretability of machine learning in different areas of healthcare, including prediction of health-related outcomes, optimizing treatments or improving the efficiency of screening for specific conditions. Further, we outline future directions for interpretable machine learning and highlight the importance of developing algorithmic solutions that can enable machine-learning driven decision making in high-stakes healthcare problems.

**Graphical/Visual Abstract and Caption**

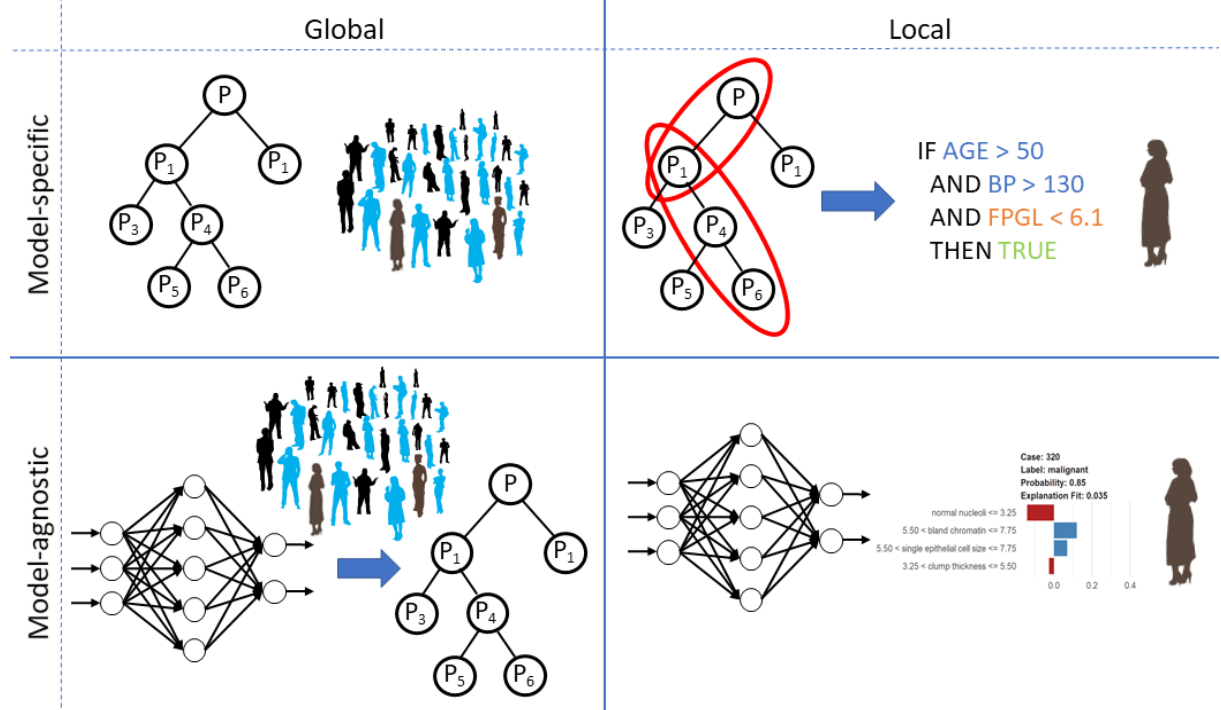

Four groups of machine learning models for prediction in healthcare based on their interpretability characteristics

**Introduction**
There is a widespread usage of artificial intelligence (AI) due to tremendous progress in technology and industrial revolution (Adadi & Berrada, 2018). The machine learning (ML) systems have shown a great success in analysis of complex patterns (Murdoch, et al., 2018) and are present in a wide range of applications in different fields, including healthcare. Thus, the need for interpretable machine learning systems has increased as well (Carvalho, Pereira, and Cardoso, 2019). Further, the lack of interpretability is a key factor that limits wider adoption of ML in healthcare. This is because healthcare workers often find it challenging to trust complex machine-learning models because they are often designed and rigorously evaluated only on specific diseases in a narrow environment and they depend on one's technical knowledge of statistics and machine learning. Bias of prediction models, for example towards black people as demonstrated in a study by Obermeyer and colleagues (2019), can further undermine the trust of the healthcare experts and other end-users of such models. Furthermore, applying those models on the larger systems may not perform well because of the complexity of the data and diversity of patients and diagnoses. Moreover, most of the models focus on accuracy prediction and rarely explain their predictions in a meaningful way (Elshawi, Al-Mallah, and Sakr, 2019; Ahmad, et al., 2018). This is especially problematic in healthcare applications, where achieving high predictive accuracy is often as important as understanding the prediction. Interpretable ML has thus emerged as an area of research that aims to design transparent and explainable models and develop means to transform black-box methods into white-box methods whose predictions are accurate and can be interpreted meaningfully. Interpretable ML is fundamentally complex and represents a rapidly developing field of research (Hall & Gill, 2018; Gilpin, et al., 2019). Several terms can be found describing the interpretability and related concepts, such as the multiplicity of good models or model locality (Breiman, 2001; Hall & Gill, 2018), comprehensibility or understandability (Piltaver, et al., 2016), and mental fit or explanatory ML (Bibal & Frenay, 2016). Interpretability is frequently defined as a degree to which a human can understand the cause of a decision from a ML model (Miller, 2017). It can also be defined as an ability to explain the model outcome in understandable terms to a human (Doshi-Velez & Kim, 2017). Gilpin and colleagues (2019) describe the main goal of interpretability as the ability to describe system internals to humans in an understandable way.

It is important that ML models are interpretable, because the higher the interpretability of the model, the easier it is to comprehend why certain predictions have been made (Molnar, 2018). Moreover, the European General Data Protection Regulation (GDPR) policy on the rights of citizens states that researchers must explain algorithmic decisions that have been made in ML models (Wachter, Mittelstadt, & Floridi, 2017; Wallace & Castro, 2018; Greene, et al., 2019). Consequently, this means that there must be a possibility to make the results re-traceable or in other words, the end-user has the right to explanation of all decisions made by the computer. Moreover, traits like fairness, privacy, reliability or robustness, causality and trust should be considered when developing ML based prediction models (Doshi-Velez & Kim, 2017; Ahmad et al. 2018). On the other hand, one can also find literature questioning the importance of the model interpretability. For example, in a recent review paper on the state of AI in healthcare Wang, and Preininger (2019) quote Geoff Hinton, one of the pioneers in the field of AI, saying: "Policymakers should not insist on ensuring people understand exactly how a given AI algorithm works, because people can't explain how they work, for most of the things they do."

In general, approaches for interpretability of ML models are either global or local. Traditionally the focus in ML research was on global interpretability in an effort to help users understand the relationship between all possible inputs to a ML model and the space of all predictions made by the model (Bratko, 1999; Martens, et al., 2008). In contrast, local interpretation helps users to understand a prediction for a specific individual or about a small, specific region of the trained prediction function (Hall, et al, 2019; Stiglic, et al., 2006). Both types of approaches have been successfully used in different healthcare domains and are still being advanced in parallel with novel methodological approaches in development of ML based prediction models. This paper outlines the basic taxonomy of ML based prediction models in healthcare and offers some examples of research studies demonstrating the use of specific approaches to interpretability in real-world studies.

It must be noted that approaches focusing on the interpretability of ML models built on non-structured data such as different types of medical images, text or other signal-based data are not included in this study. To obtain more information on a more general view of interpretability and explainable AI in healthcare, the reader is referred to works by Holzinger and colleagues (2019) or London (2019). Recently a broader concept of responsible ML in healthcare was presented (Wiens, et al., 2019), which proposes that an interdisciplinary team of engaged stakeholders (policy makers, health system leaders and individual researchers) systematically progress from problem formulation to widespread deployment, where also the level of required transparency and thus interpretability is addressed.

**Interpretability of machine learning models**
There are different criteria for classifying methods for machine learning interpretability such as intrinsic or post-hoc classification, pre-model, in-model or post-model and classification based on the model outcome (Molnar, 2018; Carvalho, Pereira & Cardoso, 2019). Intrinsic interpretability is usually described as selecting and training a machine learning model that is intrinsically interpretable due to their simple structure (e.g. simple decision tree or regression model). Post-hoc (post-model) interpretability usually describes application of interpretability methods after the training of the model (Molnar, 2018). Interpretability methods can also be classified based on the time of building the machine learning model. Pre-model methods are independent of the model and can be used before the decision on which model will be used is taken. These methods are for example descriptive statistics, data visualization, Principal Component Analysis (PCA), t-Distributed Stochastic Neighbor Embedding (t-SNE), and clustering methods. So called in-model prediction models have inherent interpretability integrated in the model itself. Post-model interpretability refers to improving interpretability after building a model (post-hoc) (Carvalho, Pereira & Cardoso, 2019). Moreover, variety of post-hoc interpretability methods provide insight into what model has learned, without changing the underlying model (Murdoch, et al., 2018).

Interpretability methods can also be classified by the outcome of the prediction model. Examples of such approaches include the following methods (Molnar, 2018):
- Feature summary statistics which refer to summary of each feature statistics that affects the model predictions.
- Feature summary visualization describing the methods that can only be visualized and could not be meaningfully presented in the form of a table.
- Model internals approach refers to the interpretation of intrinsically interpretable models and linear models, where the model weights represent both, model internals and summary statistics for the features.
- The data point interpretability refers to methods that return data points to make model interpretable. Such methods require the data points themselves to allow interpretability.
- Intrinsically interpretable models are interpretable by internal model parameter of feature summary statistics.

In the clinical setting ML prediction models are different than models in other fields, since the predictions can have an impact on patient's health, therefore the users of such models need to have confidence in the predictions of such models. We propose a categorization of interpretability that is simple and can be used by experts and non-experts in the field. Therefore, we categorize interpretability approaches in two non-exclusive groups – i.e. model-specific or model-agnostic and local or global.

**Model-specific or model-agnostic interpretability**
Model-specific interpretation methods are limited to specific models and derive explanations by examining internal model parameters (Du, Liu & Hu, 2019). Model-agnostic methods can be used on any machine learning model and are usually applied post hoc (Molnar, 2018). Most frequently, internal model parameters are not inspected as the model is treated as a black box (Du, Liu & Hu, 2019). Typically, to achieve model-agnostic interpretability, one can use a surrogate or a simple proxy model to learn a locally faithful approximation of a complex, black-box model based on outputs returned by the black-box model. This approach was introduced in a paper on model compression by Bucila, Caruana, and Niculescu-Mizil (2006) and was recently revived and updated under the name of knowledge distillation (Hinton, Vinyals & Dean, 2015).

Here we will briefly mention another type of model-specific interpretability as implemented in Graph Neural Network Explainer (GNNExplainer) (Ying, et al., 2019), which arises from a specific form of complex data representation. GNNExplainer uses graph representation where the complexity of data representation requires a special ML framework, such as the current state-of-the-art Graph Neural Network (GNN), which is experiencing a surge of interest (Xu, et al., 2019). The details are outside the purview of this article, the reader is referred to works by Hamilton and colleagues (Hamilton, Ying & Leskovec, 2017), but it should be noted that GNNExplainer provides both global and local interpretability and might prove useful for practitioners of GNN in the future, as it provides the ability to visualize relevant structures to interpretability and gives insights into errors of faulty GNNs (Ying, et al., 2019).

**Local or global interpretability**
Methods can explain a single prediction (local interpretability) or the entire model behaviour (global interpretability) (Molnar, 2018). Local interpretation of the models can be achieved by designing justified model architectures that explains why a specific decision was made. It can also be achieved by providing similar examples of instances to the target instance. For example, by emphasizing specific characteristics of a patient that are similar to characteristics of a smaller group of patients but different in other patients. On the other hand, globally interpretable models offer transparency about what is going on inside a model on an abstract level (Du, Liu & Hu, 2019). In order to explain global output of the model one needs a trained model, knowledge about the algorithm and the data (Lipton, 2016). Some authors (Ahmad, et al., 2018) also argue that models can have cohort-specific interpretability, where they focus on population subgroups. However, we can also classify such cases as either global if the subgroup is treated as the sub-population or as local if single prediction interpretations for the subgroup are grouped together (Molnar, 2018). An example of such model is Model Understanding through Subspace Explanations (MUSE), where it uses the behaviour of subspaces characterized by certain features of interest for explanation (Lakkaraju, et al., 2019).

**Examples of interpretable machine learning models**
Some examples of prediction models that are interpretable using global/local or model-specific/model-agnostic interpretability techniques are shown in Table 1.

**Table 1.** Examples of approaches to interpretability of prediction models

|  | **Global** | **Local** |
|---|---|---|
| **Model-specific** | - Decision trees,<br>- Regression models,<br>- Naive Bayes classifier,<br>- GNNExplainer (Ying, et al., 2019) | - Set of rules (for specific individual),<br>- decision trees (by tree - decomposition),<br>- most visual analytics-based approaches,<br>- k-nearest neighbours,<br>- GNNExplainer (Ying, et al., 2019) |
| **Model-agnostic** | - Different variants of model compression / knowledge distillation / global surrogate models,<br>- Partial Dependence Plots (PDP),<br>- Individual Conditional Expectation (ICE) plots,<br>- Black Box Explanations through Transparent Approximations (BETA) (Lakkaraju, et al., 2017)<br>- Model Understanding through Subspace Explanations (MUSE) (Lakkaraju, et al., 2019). | - Local interpretable model-agnostic explanations (LIME) (Ribeiro, et al., 2016),<br>- Shapley additive explanations (SHAP) (Lundberg & Lee, 2017),<br>- Anchors (Ribeiro, et al., 2018),<br>- attention map visualization,<br>- Model Understanding through Subspace Explanations (MUSE) (Lakkaraju, et al., 2019). |

Based on the examples given in Table 1, we can characterize some most widely used approaches that are summarized in Figure 1. Historically, global interpretability was widely used to extract knowledge from the prediction models. Model-agnostic and model-specific models are used to either interpret the decision directly from the model (e.g. extracting all rules from the decision tree) or use some specific technique like knowledge distillation (Hinton, Vinyals & Dean, 2015) to build a simple model that can be interpreted (e.g. a decision tree, set of rules or a regression function). On the other hand, local interpretation techniques were not so frequently used until recently. There are some exceptions, especially in some of the local model-specific approaches like extracting a limited number of rules from a decision tree that correspond to a specific individual. On the other hand, there were many novel techniques introduced in the last ten years that allow at least feature importance estimation for prediction at the personalized level suitable for the models with no or weak interpretability (Lundberg & Lee, 2017; Ribeiro, et al., 2016; Ribeiro, et al., 2018).

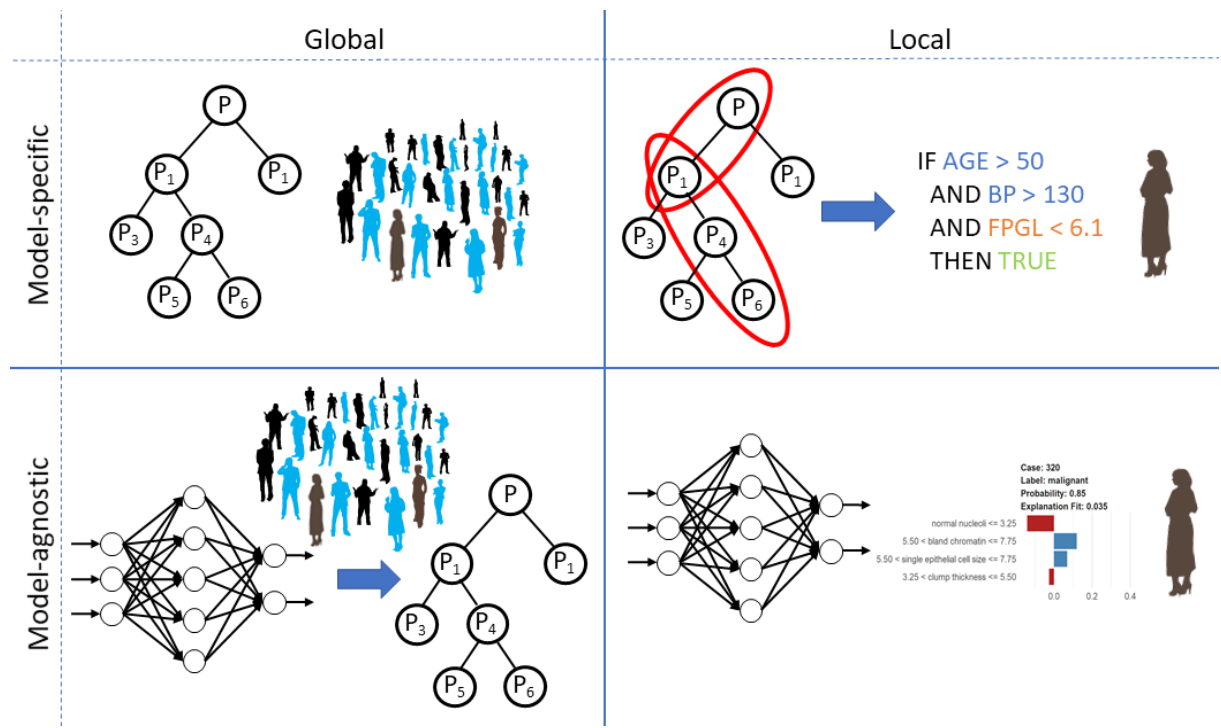

**Figure 1.** Visual summary of the most representative interpretability approaches for machine learning based predictive modelling in healthcare

**Applications of interpretable machine learning in healthcare**
In this section we provide examples of real-world studies that employed at least one interpretability approach mentioned in the previous section to demonstrate their use in different fields of healthcare.

Global model-specific approaches to interpretability of ML based models in healthcare have been in use for more than two decades. Due to their high level of interpretability and simple use in practice, the approaches like Regression models or Naive Bayes classifier are still used in different fields of healthcare like urology (Zhang, et al., 2019; Otunaiya & Muhammad, 2019), toxicology (Zhang, et al., 2018, Zhang, et al., 2019), endocrinology (Alaoui, et al. 2019), neurology (Zhang & Ma., 2019), cardiology (Joshi, et al., 2019; Salman, 2019; Feeny, et al., 2019) or psychiatry (Guimarães, et al., 2019; Obeid, et al., 2019). Model-specific approaches focusing on local interpretation that can be based for example on KNN or decision trees were recently used for interpretation in prediction of health-related conditions, including occupational diseases (Di Noia, et al., 2019) or knee osteoarthritis (Jamshidi, et al., 2019), cancer (e.g. breast cancer or prostate cancer) (Oladele Aro, et al., 2019; Seker, et al., 2019), severity of a disease, including chronical diseases (e.g. diabetes or Alzheimer's disease) (Karun, et. al., 2019; Bucholc, et al., 2019) and mortality rates (e.g. myocardial infarction or perinatal stroke) (Prabhakararao & Dandapat, 2019; Gao, et al., 2019). Local and model-agnostic interpretability can be used in interpretability of complex models, such as deep learning models. For example, SHAP was used in interpretation of predictions for the prevention of hypoxaemia during surgery (Lundberg, et al., 2018), which increased the anaesthesiologists anticipation of hypoxaemia events by 15%, or in the detection of acute intracranial haemorrhage from images of CT scans (Lee, et al., 2019), which was achieved via attention maps visualization, where areas were highlighted (segmentation map). The highlighted results corresponded to the classification and could be understood by a practitioner (Lee, et al., 2019), but it should be noted that such approaches are not scalable.

As already mentioned in the previous section, some recent approaches to interpretability cannot be simply classified in one of the four groups displayed in Figure 1. Model interpretability focusing on feature subspaces defined by the domain experts was proposed by Lakkaraju, et al. (2019). MUSE was used to help in explaining decisions from the 3-level neural network trained on depression diagnosis dataset. It generates sets of if-then rules that describe the model decisions on a global level, but it also provides a separate set of rules for a subspace based on the features selected by a healthcare expert working with patients. More specifically, authors demonstrate the effectiveness of MUSE to produce rules optimised for fidelity, unambiguity, and interpretability for a subspace focusing on excursive and smoking as those two features might be chosen by the expert as actionable features. A typical global interpretability approach would not consider the fact that there are features in the datasets that are more interesting than others as it is possible for the patient and healthcare expert to influence their values by introducing some interventions. As such, MUSE could be classified as a global model-agnostic interpretability approach, but it also demonstrates characteristics of approaches focusing on personalised interpretation by narrowing the subspace of search by end-user input.

**Discussion**

This paper provides a current and practical overview of interpretability methods for prediction models in healthcare, with a focus on usability from an end-user perspective. In contrast to many other fields, decisions in healthcare are high-stakes decisions as they can directly influence a treatment outcome or even survival of a patient. In addition to technical challenges related to the development of interpretable models, we also need to address a myriad of ethical, legal, and regulatory challenges, e.g., the GDPR's right to explanation (Wachter, Mittelstadt, & Floridi, 2017; Wallace & Castro, 2018). We categorise interpretability both in terms of model-specific and model-agnostic or global and local interpretability. The latter type of interpretability is usually represented as a user-centric approach that enables users to find an appropriate model for a specific problem. For example, when the interpretation of general predictions at the population level is required, then global and model-specific models might be an appropriate choice. In recent years, a shift can be observed from model-specific and global interpretable models to model-agnostic and local interpretable models, where one of the reasons is the availability of massive datasets in healthcare (Esteva, et al., 2019) and consequently a shift to precision medicine.

It needs to be noted that this overview does not provide an in-depth analysis of some topics that are still open (Ahmad, et al., 2018), such as simplification of complex models for explanations that may result in suboptimal results, thus auditing output for fairness and bias might be a better solution (Tomasello, 1999), scalability of interpretable ML models (e.g. LIME or SHAP values can be computationally expensive) and methods for evaluation of explanation models (e.g. different models with similar error rate but different explanations). Another current research topic related to interpretability of ML models in healthcare is causality (Holzinger, et al., 2019), which is defined as causal understanding with effectiveness, efficiency, and satisfaction in a specified context of use.

> **Sidebar title: Visual analytics and interpretability**
>
> In recent years, researchers are increasingly relying on Visual Analytics (VA) techniques to support interpretability of machine learning models within healthcare fields (Simpao et al. 2014). VA is the science of analytical reasoning facilitated by interactive visual interfaces (Thomas & Cook, 2005). In order to facilitate interpretation of complex data and models, VA techniques combine concepts from data mining, machine learning, human computing interaction, and human cognition (Caban & Gotz, 2015). Visual analytics extends the interaction paradigm of traditional information visualisation techniques to support both the interpretation of machine learning models and model steering with feedback from end-users (Endert et al. 2017). In contrast to 'black box'-driven approaches, the objective is to allow domain expert users to interpret and diagnose models.
>
> Both model-specific and model-agnostic approaches have been researched in VA prototypes. RetainVis (Kwon et al., 2018) is a prominent example of a model-specific VA approach that visualizes a recurrent neural network to support interpretation and diagnosis of the model. RetainVis uses t-SNE (Maaten & Hinton, 2008) to project patients on a 2D space and explains the model's interpretation of data by showing which patients are closely located in the latent space. The approach has been used to predict future diagnosis of heart failure and cataract. RuleMatrix (Ming, Qu, & Bertin, 2018) is a model-agnostic example that relies on extracted rule-based knowledge from the input-output behavior of a model. The system helps domain experts understand and inspect classification models using rule-based explanations and was used to improve cancer and diabetes classification.

**Conclusion**

Understanding why machine learning models reach specific outcomes is important for both model developers and end-users. In order to trust a specific model and its predictions, it is important that we understand different approaches of model interpretability. Interpretability can be categorised in terms of model-specific and model-agnostic or global and local interpretability. There are various new techniques and approaches available for interpretable machine learning. However, the key challenges are still unsolved, and future research is needed to find new and reasonable solutions for the progress in this field.

In the future, we expect more approaches like MUSE (Lakkaraju, et al., 2019) where a global interpretability approach is supplemented by specific interpretations that can explain either a single individual or a smaller group of individuals. Although it is difficult to predict the exact direction of future research in this field, it is certain that interpretability techniques represent an important concept that needs to be taken into account when developing prediction models for healthcare. Another avenue of research might lie with GNN, which combines node feature information and graph structure by using neural networks (Ying, et al., 2017) and the aforementioned GNN Explainer as a tool for post-hoc explanation of GNN. However, there is a limited understanding of GNNs properties and limitations (Xu, et al., 2019). Moreover, it is necessary to understand and additionally improve the generalization properties of GNNs. Future research should focus on developing algorithmic solutions that enable machine-learning driven decision-making for various healthcare problems that influence disease course and outcome.

**Funding Information**
This work was supported by the Slovenian Research Agency grants ARRS N2-0101 and ARRS P2-0057.